\documentclass{article}

\usepackage{arxiv}

\usepackage[utf8]{inputenc} 
\usepackage[T1]{fontenc}
\usepackage{hyperref} 
\usepackage{url} 
\usepackage{booktabs} 
\usepackage{amsfonts} 
\usepackage{nicefrac} 
\usepackage{microtype}
\usepackage{tabulary}
\usepackage{amsfonts,amsmath,amssymb}
\usepackage{graphicx}
\usepackage[table,xcdraw]{xcolor}
\usepackage[para,online,flushleft]{threeparttable}
\usepackage{multirow}
\usepackage{tikz}
\usetikzlibrary{shapes,decorations,arrows,calc,arrows.meta,fit,positioning,decorations.pathreplacing,trees}
\tikzstyle{level 1}=[level distance=3cm, sibling distance=3cm, font=\scriptsize]
\tikzstyle{level 2}=[level distance=3cm, sibling distance=1.5cm,font=\scriptsize]
\tikzstyle{bag} = [text width=4em, text centered, font=\scriptsize]
\tikzstyle{end} = [circle, minimum width=2pt, fill, inner sep=0pt]
\tikzset{
    -Latex,auto,node distance =1 cm and 1 cm,semithick,
    state/.style ={ellipse, draw, minimum width = 0.7 cm},
    point/.style = {circle, draw, inner sep=0.04cm,fill,node contents={}},
    bidirected/.style={Latex-Latex,dashed},
    el/.style = {inner sep=2pt, align=left, sloped}
}

\title{Causal inference for observational longitudinal studies using deep survival models}

\author{
Jie Zhu\thanks{Corresponding Author} \\
 Centre for Big Data Research in Health (CBDRH)\\ UNSW, Sydney\\ NSW, 2052, Australia\\
\texttt{elliott.zhu@unsw.edu.au} \\
 \And
Blanca Gallego \\
 Centre for Big Data Research in Health (CBDRH)\\ UNSW, Sydney\\ NSW, 2052, Australia\\
\texttt{b.gallego@unsw.edu.au} \\
}

\begin{document}
\maketitle

\clearpage

\begin{abstract}
\textbf{Objective}
Causal inference for observational longitudinal studies often requires the accurate estimation of treatment effects on time-to-event outcomes in the presence of time-dependent patient history and time-dependent covariates. 

\textbf{Materials and Methods}
To tackle this longitudinal treatment effect estimation problem, we have developed a time-variant causal survival (TCS) model that uses the potential outcomes framework with an ensemble of recurrent subnetworks to estimate the difference in survival probabilities and its confidence interval over time as a function of time-dependent covariates and treatments. 

\textbf{Results}
Using simulated survival datasets, the TCS model showed good causal effect estimation performance across scenarios of varying sample dimensions, event rates, confounding and overlapping. However, increasing the sample size was not effective in alleviating the adverse impact of a high level of confounding. In a large clinical cohort study, TCS identified the expected conditional average treatment effect and detected individual treatment effect heterogeneity over time. TCS provides an efficient way to estimate and update individualized treatment effects over time, in order to improve clinical decisions.

\textbf{Discussion}
The use of a propensity score layer and potential outcome subnetworks helps correcting for selection bias. However, the proposed model is limited in its ability to correct the bias from unmeasured confounding, and more extensive testing of TCS under extreme scenarios such as low overlapping and the presence of unmeasured confounders is desired and left for future work. 

\textbf{Conclusion}
TCS fills the gap in causal inference using deep learning techniques in survival analysis. It considers time-varying confounders and treatment options. Its treatment effect estimation can be easily compared with the conventional literature, which uses relative measures of treatment effect. We expect TCS will be particularly useful for identifying and quantifying treatment effect heterogeneity over time under the ever complex observational health care environment.

\end{abstract}


\clearpage
\section{Introduction}
While randomized experiments are the gold standard in the comparison of interventions, it has become clear that observational studies using Big Data have an important role to play in comparative effectiveness research \cite{793296:20590042}. As a result, the last few years have seen a surge of studies proposing and comparing methods that can estimate the effect of interventions from routinely collected data. In particular, new methods have emerged that can investigate the heterogeneity of the treatment effect. In the health domain, these methods use Medical Claims \cite{793296:20590051} and Electronic Health Record (EHR) data and have been driven by the move towards personalized care \cite{793296:20590055}.

In spite of significant progress, there remain challenges that must be addressed. In particular, no off-the-shelf treatment effect algorithm exists that takes full consideration of the temporal nature of medical data such as the time-dependent patient history and time-to-event outcomes. Accounting for the temporal nature of medical information is important when informing clinical guidelines or designing clinical decision support systems, since they underpin clinicians' response to disease progression and patient deterioration.

As a motivating example, early detection and treatment of sepsis are critical for improving sepsis outcomes, where each hour of delayed treatment has been associated with roughly a 4-8\% increase in mortality \cite{793296:20397422}. To address this problem, clinicians have proposed new definitions for sepsis \cite{793296:20397467}, but the fundamental need to detect and treat sepsis early still remains. In this context, time-dependent variables such as previous administration of antibiotics or use of mechanical ventilation (MV) may play a significant role on treatment decisions and their corresponding outcomes. The challenge of capturing the history of time-dependent biomarkers and other risk factors pervades the prediction of time-to-event outcomes and the estimation of their treatment effects.

The standard method to estimate the treatment effect using time-dependent confounders uses the Cox model \cite{793296:19632802} such as in the landmark analysis \cite{793296:19639435}, where the instantaneous probability of experiencing an event at time $t $ given covariates $X(t)$ is defined as a hazard function: $h(t|X(t)) = h_0 \exp( \beta' X(t)) $, where $\beta $ is a vector of constants, and $h_0$ is the baseline risk of having an event at time 0. When censoring is not considered, a Cox model compares the risk of an event between treatment and control conditions at each time $t$ regardless of previous history of $X(t) $ or history of treatment conditions.The piece-wise constant Cox model \cite{793296:19693408} extends the constant$\beta $ to $\beta(t)$ thus allowing for a time-dependent effect. However, neither model takes into account the longitudinal history of covariates and both treat missing covariates either by imputing their value or removing the incomplete observations.

To address these limitations, models were proposed to jointly describe both longitudinal and survival processes \cite{793296:19639042,793296:19639112}. In particular, these joint models generally comprise two submodels: one for repeated measurements of time-dependent covariates and the other for time-to-event data such as a Cox model. The models are linked by a function of shared random effects. To find a full representation of the joint distribution of the two processes, the model needs to be correctly specified for both processes. Thus, model misspecification and computation efforts significantly limit the estimation accuracy of this approach when applied to high-dimensional EHR data.

Recently, data-driven models such as recurrent neural networks \cite{793296:19702611,793296:19504300} have been proposed to learn efficiently from EHR data with complex longitudinal dependencies. For example, Dynamic DeepHit \cite{793296:19504300} is a longitudinal outcome model which learns the joint distribution of survival times and competing events from a sequence of longitudinal measurements with a recurrent neural network structure. However, as a single outcome prediction model, DeepHit does not provide an explanatory mechanism for causal inference.

On the other hand, the recently proposed Counterfactual Recurrent Network (CRN) \cite{793296:19702611} estimates the average longitudinal treatment effect on continuous outcomes by correcting for time-dependent confounding using domain adversarial training (DAT). However, the efficacy of DAT depends on the feature alignment in the source (control) and target (intervention) domains \cite{793296:19783373}, that is, whether the covariates observed under treatment and control conditions have similar distributions. As shown in the original work, the DAT is sensitive to the overlapping among covariates and drops in estimation performance with the level of overlapping like other existing causal inference algorithms such as the TMLE \cite{793296:19514627} and Causal Forest \cite{793296:20677743}. 

There is a lack of studies dedicated to the estimation of survival causal effects from longitudinal EHR data. We fill this gap by introducing a time-variant causal model for survival analysis, which we call TCS, which extends our previous work on modeling the treatment effect on time-to-event outcomes from static patient history \cite{zhu2020targeted}. In TCS, we choose an ensemble of recurrent neural networks as the outcome model. Neural networks learn more efficiently from the trajectories of covariates than semi-parametric or parametric methods such as Super-learner and Cox models, while the ensemble captures the uncertainty of network estimation. Both the baseline survival probability and its interaction with treatment can vary with time free from the proportional hazard assumption.

In lieu of a single outcome model (like DeepHit) for estimating the joint distribution of the observed failure/censor times, TCS first captures the information from treated and control observations separately, and then encodes it into a shared subnetwork. The encoded information is fed into counterfactual subnetworks to predict the expected survival outcomes given either treatment or control conditions. The dedicated subnetworks explicitly model the outcomes originated from patient baseline covariates and their interaction with treatment conditions. Lastly, we adjust for bias in the counterfactual outcomes arising from nonrandom treatment allocation in observational studies. The difference between the counterfactual survival probabilities will give us adjusted treatment effect estimates.

The key characteristics of the proposed algorithm are: 1) it learns the treatment assignment and outcome generating processes from the pattern of observed and missing covariates in longitudinal data; 2) it captures treatment specific outcomes by employing potential outcome subnetworks for treatment and control conditions; 3) it quantifies the uncertainty of the model estimations with an ensemble of neural networks with varied random seeds; and 4) it incorporates the history of previous treatments as additional covariates, allowing for straightforward updating of treatment effect estimations over time.

The outline of this project is as follows. Section 2 describes the materials and methods. Section 3 provides the results. We end with a discussion.

\section{Materials and Methods}

\subsection{The case study}

TCS provides a solution to the need to analyze the high-dimension time-dependent observations in the patient history. It predicts patient outcomes in terms of survival probabilities from time-dependent patient history without feature engineering and estimates conditional treatment effects for selected patient groups. 

We illustrate the TCS model in evaluating the effectiveness of mechanical ventilation (MV) on in-hospital mortality for sepsis patients in the ICU. The data source for this case study is MIMIC-III, an open-access, anonymised database of 61,532 admissions from 2001{\textendash}2012 in six ICUs at a Boston teaching hospital \cite{793296:20320029}.

\begin{table*}[!htbp]
\footnotesize
\caption{Summary of case study database}
\label{tw-f196b3db0db7}
\centering 
\begin{tabular}{lll}
&\textbf{MIMIC-III} \\
\hline 
Unique patient ids & 20,938 \\
Number of event patients & 2,880 \\
Rows for the first 20 time stamp & 278,504 \\
Static features & 5 \\
Dynamic features & 39 \\
\hline 
\end{tabular}
\end{table*}

In our case study, we define a sepsis patient as those who had a record of suspected infection (identified by the prescription of antibiotics and sampling of bodily fluids for microbiological culture) and the evidence of organ dysfunction (defined by a two-points deterioration of the SOFA score \cite{793296:20396889}). The final cohort has 20,938 patients (including both adults and non-adults, please see Figure 1 in the supplementary information of the previous work for detail \cite{793296:20397006}) and its summary is presented in Table~\ref{tw-f196b3db0db7}. We consider the first 20 timestamps\footnote{The first 20 2-hour intervals for MIMIC-III, the discretization process takes the average value of each covariate during the interval and the missing values are masked as described in the method section.} of each patient for the treatment effect estimation. 

The treatment, which is the use of mechanical ventilation(MV), is a time-invariant binary covariate. However, as illustrated in Table~\ref{tw-f196b3db0db7}, there are 39 time-variant covariates making the adjustment for treatment effect a challenging task. In our simulation study, we further allow the treatment to be time-variant in order to generalize the application of the model. We train the model using the 10-fold cross validation with 70\% of the original data injected in each training epoch. We estimate the average treatment effect at each time step using the whole sample.

\subsection{A causal model for time-variant survival analysis}

As illustrated in Figure \ref{f-Dis51}, suppose we observe a sample $\mathcal{O} \text{ of } N$ independent observations generated from an unknown distribution $\mathcal{P}_0 $:

\begin{figure*}[!htbp]
\centering 
\includegraphics[width = \linewidth]{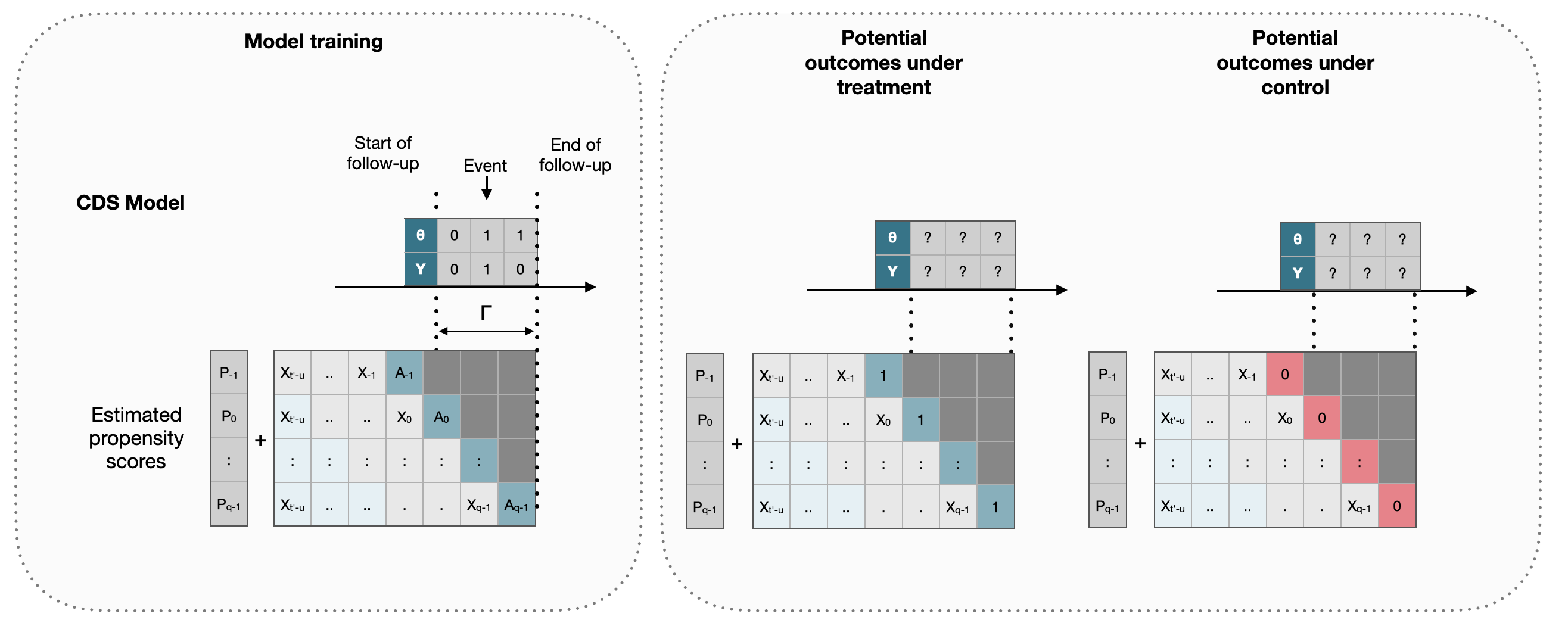}
\caption{Treatment effect estimation with time-varying covariates using potential outcomes. All dark gray cells indicate missing data. The TCS model estimates treatment effects as a function of time-dependent covariates, which is defined as the difference between potential survival probabilities from $t=0$ (start of follow-up) to $t=q$ (end of follow-up) given constant treatment or control conditions. }
\label{f-Dis51}
\end{figure*}

\begin{eqnarray*}
\mathcal{O} := \big(X_i(t), Y_i(t), A_i(t), \tau_{i} = \min(\tau_{s,i},\tau_{c,i})\big),i~= 1,2,\ldots, n \end{eqnarray*}
\noindent where $X_i(t) = (X_{i,1}(t), X_{i,2}(t),\ldots,X_{i,\textit{d}}(t)), d=1,2, \ldots,D$ are covariates at time $t$; $A_i(t)$ is the treatment condition at time $t$, which can take the value of $0$ or $1$ for control and treatment conditions respectively; and $Y_i(t)$ denotes the outcome at time $t$, with $Y_i=1 $ if $i$ experienced an event and $Y_i=0$ otherwise. Both $X_i(t)$ and $A_i(t)$ are captured from $t=-u$ to $t=q$ (inclusive),$u>1$ and $q>1$, where $u$ is the length of the patient's history window. The end of follow-up for a given patient, $\tau_{i}, \tau_{i} \geq 0$, is determined by the event or censor time, $\tau_{s,i}$ or $\tau_{c,i}$, whichever happened first. For simplicity, we drop the subscript $i$ in the sequel.

To fit the TCS model and adjust for right-censoring, we create the longitudinal outcome label, $\Gamma$, as a matrix: 

\begin{align}
\begin{split}
\label{dfg-364b5b929424}
\Gamma&=\begin{bmatrix}\Theta\\ \gamma \end{bmatrix},\\
\end{split}
\end{align}

\noindent composed of the vector of events $\gamma = [Y(1),\dots,Y(\tau),\ldots,Y(q)]$ and the vector of terminal timing labels $\Theta= [\theta(1),\dots,\theta(\tau),\ldots,\theta(q)]$, where $\theta(\cdot) = 1$ for  $t<\tau$ if a patient is censored or has an event at $\tau$, and $\theta(\cdot) = 0 $ for $t\geq \tau$. As shown in Appendix A, the estimation $\hat\theta(t)$ using TCS is equivalent to the hazard rate of experiencing an event at time $t$ adjusted for right-censoring. 

\begin{figure*}[!htbp]
\centering 
\includegraphics[width = 0.5\linewidth]{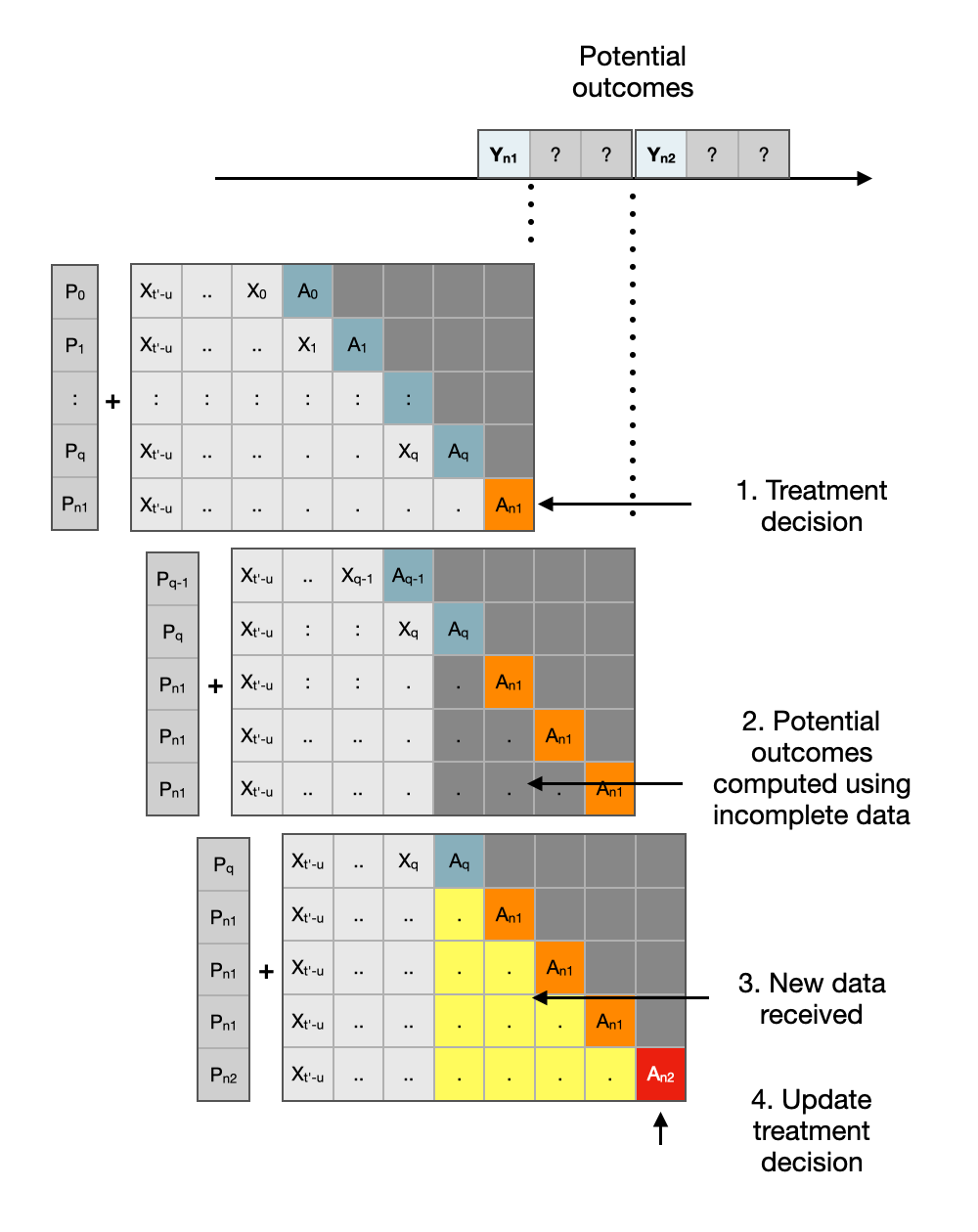}
\caption{Illustration of treatment recommendations based on potential outcomes. All dark gray cells indicate missing data. At time $n_1$, a treatment option, $A_{n_1}$, is selected and its corresponding potential outcome $Y_{n_1}$ is computed (with right-censoring and confounding adjusted) . Practitioner may decide whether $A_{n_1}$ is appropriate given $Y_{n_1}$. When there are multiple time steps, the decision process can be repeated as illustrated in steps 3 and 4. }
\label{f-Dis52}
\end{figure*}

TCS maps the propensity score and covariate matrices in Figure \ref{f-Dis51} (which we denote as $\Lambda$) to the outcome $\Gamma$:

\begin{align}
 \Gamma = f(\Lambda)
\end{align}

The potential survival curves for the treatment arm are computed mapping $\hat\Gamma_1 = f(\Lambda_1)$, where $\Lambda_1$ is calculated by setting all $A(t)=1$ in $\Lambda$. Similarly, the potential survival curves under the control condition are computed mapping $\hat\Gamma_0 = f(\Lambda_0)$, where where $\Lambda_0$ is calculated setting all $A(t)=0$. This setup accounts for time-dependent covariates throughout the follow-up window, but assumes that there is no treatment-covariate feedback, that is the covariates observed after the treatment assignment during the follow-up are independent from previous treatments. 

We use the conditional probability,

\begin{align}
\begin{split}
\label{dfg-2c1cf113437b}
S(t,\Lambda) &= \prod_{j=1}^{t}(1-\hat\theta(j,\Lambda)),
\end{split}
\end{align}

\noindent to denote the time-to-event probabilities that the event did not occur in any observation from time 1 to time $t$ conditioned on $\Lambda$.TCS can be easily extended to a system that provides treatment recommendations at selected times based on estimated potential outcomes as illustrated in Figure \ref{f-Dis52}. We discuss the implementation of TCS in Appendix A.

\subsection{Define the survival treatment effect}

To estimate the treatment effect over the follow-up window, we follow Rosenbaum and Rubin's potential outcomes framework \cite{793296:19013328}, and assume 1) the censoring is non-informative conditioned on the treatment (coarsening at random), 2) there is no unmeasured confounders, 3) the history of treatment assignment $\overline{A}(u)$ is independent of the outcomes given the history of correctly estimated propensity scores  $\overline{P}(u)$ and 4) $X(\tau)$ is independent of $A(t)$ for all $\tau>t$. Then the conditional average treatment effect (CATE) can be defined as:
\begin{equation}
\label{dfg-d220898f21d7}
\begin{split}
\Psi(t,\lambda)= \mathbb{E}_{\Lambda=\lambda}\big[\mathbb{E}[S(t,\Lambda_1)]-\mathbb{E}[S(t,\Lambda_0)]\big].
\end{split}
\end{equation}

\noindent Similarly, we define the individual treatment effect (ITE) as:

\begin{equation}
\label{dfg-ebbddc3a09a3}
\begin{split} 
\psi(t_i,\lambda_i)= \mathbb{E}[S(t_i,\lambda_{i,1})]- \mathbb{E}[S(t_i,\lambda_{i,0})].
\end{split}
\end{equation}

To compare the absolute measure of treatment effect with the conventionally reported hazard ratio, we define an empirical hazard ratio as: 

\begin{equation}
\label{dfg-36808aa93027}
\begin{split}
\text{HR}^*(t,\lambda)=\frac1n\sum_i\frac{\hat{\theta}_i(t, \Lambda_0)}{\hat\theta_i(t, \Lambda_1) )}
\end{split}
\end{equation}

\noindent where $n$ is the number of observations in a sample where $\Lambda=\lambda$.

\subsection{Model evaluation}

\subsubsection{Benchmark data}

To explore the finite-sample performance of TCS, we ran several experiments with biologically plausible longitudinal data following a previous study \cite{793296:19027107}. In particular, we use:

\begin{itemize}
\item $D$ continuous confounders $X(t)_1,X(t)_2,\dots,X(t)_D\sim\mathrm{N}(\sqrt{t},V)$ from $t=t'-u$ to $t=t'+q$, where $V$ is the variance of the normal distribution and $D$ is the feature dimension;

\item Binary exposure: $A(t) \sim Binom( \eta \cdot I(\sum_{d=1}^{3}X(t)_D > \frac{1}{3}\sum_{i=1}^{3}X(t)_i) + 0.5\cdot(1-\eta))$, where $I$ is an indicator function and $\eta$ controls the level of overlapping. When $\eta=0 $, the probability of receiving the treatment is 50\% regardless of $X(t)$; when $\eta=1 $, the allocation follows the indicator function so that the outcome will be confounded by the first 3 confounders of $X(t)$; and when $\eta=0.5$, the chance of receiving the treatment is partially dependent on the indicator function which is $(0.5 \cdot I~+ 0.25)100\% $.
\item Hazard rate:$h(t) = \frac{\log(t)}{\lambda}\big(0.1 A(t)+\beta\sum_{i=1}^{D} X(t) \big) $ where $\beta = 1$;
\item The survival probability $S(t) = \exp(- h(t)) $;
\item The censoring probability $SC(t)=\exp(-\frac{\log(t)}\lambda),$ where $\lambda=30 $; 
\item An event indicator generated using \textit{root-finding} \cite{793296:19027107} at each time $t$: $E(t)=I(S(t)<U\sim\mathrm{Uniform}(0,1))$, with the event time defined by $\tau_s\text{ if } E_i(\tau_s)=1$, otherwise $\tau_s = q+1 $;
\item A censoring indicator generated using the \textit{root-finding} technique: $C(t)=I(SC(t)<U\sim\mathrm{Uniform}(0,1))$, with the censoring time defined by $\tau_c\text{ if }C(\tau_c)=1$, otherwise $\tau_c = q$, and;
\item The survival outcome given by the indicator function: $Y=I(\tau_s \leq \tau_c)$.
\end{itemize}
A series of experiments were conducted by changing the following parameters: $V\in\{0.5,1.0,1.5,2.0\} $, $D\in\{6,10,20,40\} $, $\eta\in\{0.7,0.8,0.9,1.0\} $, $N \in \{1500, 3000, 10000\} $. We define our default data generation model with $V=0.5, D=6,\eta=0.9$, and $N=1500$. In this study, we set the length of the estimation window from $t'$ to $q$ at 10 time steps and the length of history window at five time steps ($u=5$). For each scenario, we generate 50 sets of training and testing samples using the same parameters but different random seeds. All evaluations are averaged over the testing results from these 50 samples. 

\subsubsection{Benchmark metrics}

The explanatory performance of TCS is assessed with simulation studies using the three metrics described below:

\textbf{Root-mean-square error (RMSE)}: Refers to the expected mean squared error of the estimated individual treatment effect: 
\begin{eqnarray*}\text{RMSE}(t) = \frac{1}{ n_k } \sum_{i_k} (\hat{\psi}_{i_k}(t) - \psi_{i_k}(t))^{2} \end{eqnarray*}
where $n_k $ is the number of individuals in subgroup $k $and $i_k $ is the individual indicator in each group. When estimating the ATE, we will have $n_k = N $, the sample size.

\textbf{Absolute percentage bias (Bias): }Defined as the absolute percentage bias in the estimated conditional average/individual treatment effect: 
\begin{eqnarray*}\text{Bias}(t) =\frac{1}{ n_k}\sum_{i_k} \big|\frac{\hat{\psi}_{i_k}(t) - \psi_{i_k}(t)}{\psi_{i_k}(t)}\big|
 \end{eqnarray*}
\textbf{Coverage ratio:} Refer to the percentage of times that the true treatment effect lies within the 95\% confidence intervals of the posterior distribution of the estimated individual treatment effect.
\begin{eqnarray*}\text{Coverage}(t)=\frac{1}{n_k}\sum_{i_k} I(|\hat{\psi}_{i_k}(t) - \psi_{i_k}(t)| < CI_{i_k}(t)) \end{eqnarray*}
where $I $ is an indicator function, $I = 1 $ if $I(\cdot) $ is true and 0 otherwise. CI is the 95\% confidence interval of the estimations.

\textbf{Concordance and AUROC: } We evaluate the models' discrimination performance of the estimated survival curves with Harrell's Concordance-index \cite{793296:20320158} and the area under the receiver operating characteristic curve (AUROC). 

\subsubsection{Benchmark algorithms}
The TCS model was benchmarked against two other machine learning algorithms:
\begin{enumerate}
\item Plain recurrent neural network with survival outcomes (SNN): this is achieved by removing the propensity score estimation layer in Figure~\ref{f-aa387bd6e319}.
\item Plain recurrent neural network with binary outcomes (Binary): direct prediction of the longitudinal outcome defined by the independent Binary labels in the first part of Equation~(\ref{dfg-364b5b929424}) using mean squared error as the loss function. 
\end{enumerate}
 
For a fair comparison, we applied the inverse probability weighting (IPW) and the iterative targeted maximum likelihood estimation (TMLE) to the raw estimations from SNN and Binary to correct for selection bias when estimating the CATE (please refer to Appendix A for a detailed explanation). We developed TCS using Python 3.8.0 with Tensorflow 2.5.0\cite{793296:20322619} (code available at \url{https://github.com/EliotZhu/TCS}).

\section{Results}

\subsection{Experiments}

In Table~\ref{tab:my-table}, we compare TCS against the selected benchmark models using the test data generated under the default scenario. The Binary method achieves the highest AUROC, while TCS and SNN models have better performance in concordance due to the survival outcomes design. In terms of treatment effect estimation, TCS achieves nominal performance in both ITE and ATE estimations compared to both IPW and TMLE adjusted ATE estimations provided by the Binary and SNN models.

\begin{table}[!hb]
\centering
\begin{threeparttable}
\footnotesize
\caption{Estimation performance by benchmark algorithms under the default scenario}
\label{tab:my-table}
\begin{tabular}{lccc}
 & \multicolumn{3}{c}{\textbf{Algorithms}} \\
\hline
\textbf{Metrics} & \textbf{Binary} & \textbf{TCS} & \textbf{SNN} \\
\hline
\textbf{AUROC} & \textbf{0.96 (0.816,1.106)} & 0.82 (0.729,0.914) & 0.85 (0.725,0.983) \\
\textbf{Concordance} & 0.76 (0.730,0.799) & \textbf{0.90 (0.852,0.947)} & 0.86 (0.798,0.913) \\
\hline
\textbf{Bias (IPW)} & 0.65 (0.625,0.675) & \textbf{-} & 0.15 (0.133,0.167) \\
\textbf{Bias (TMLE)} & 0.63 (0.576,0.684) & \textbf{-} & 0.14 (0.129,0.151) \\
\textbf{Bias (ATE)} & - & \textbf{0.10 (0.061,0.136)} & - \\
\textbf{Bias (ITE)} & 0.75 (0.706,0.794) & \textbf{0.10 (0.064,0.137)} & 0.43 (0.401,0.459)\\
\bottomrule
\end{tabular}%
\begin{tablenotes}
\footnotesize 
\item{All metrics are averaged across 50 independent simulations over 30 time points from the test dataset under the default scenario.}
\end{tablenotes}
\end{threeparttable}
\end{table}

\begin{figure*}[!htbp]
\centering {\includegraphics[width = 0.45\linewidth] {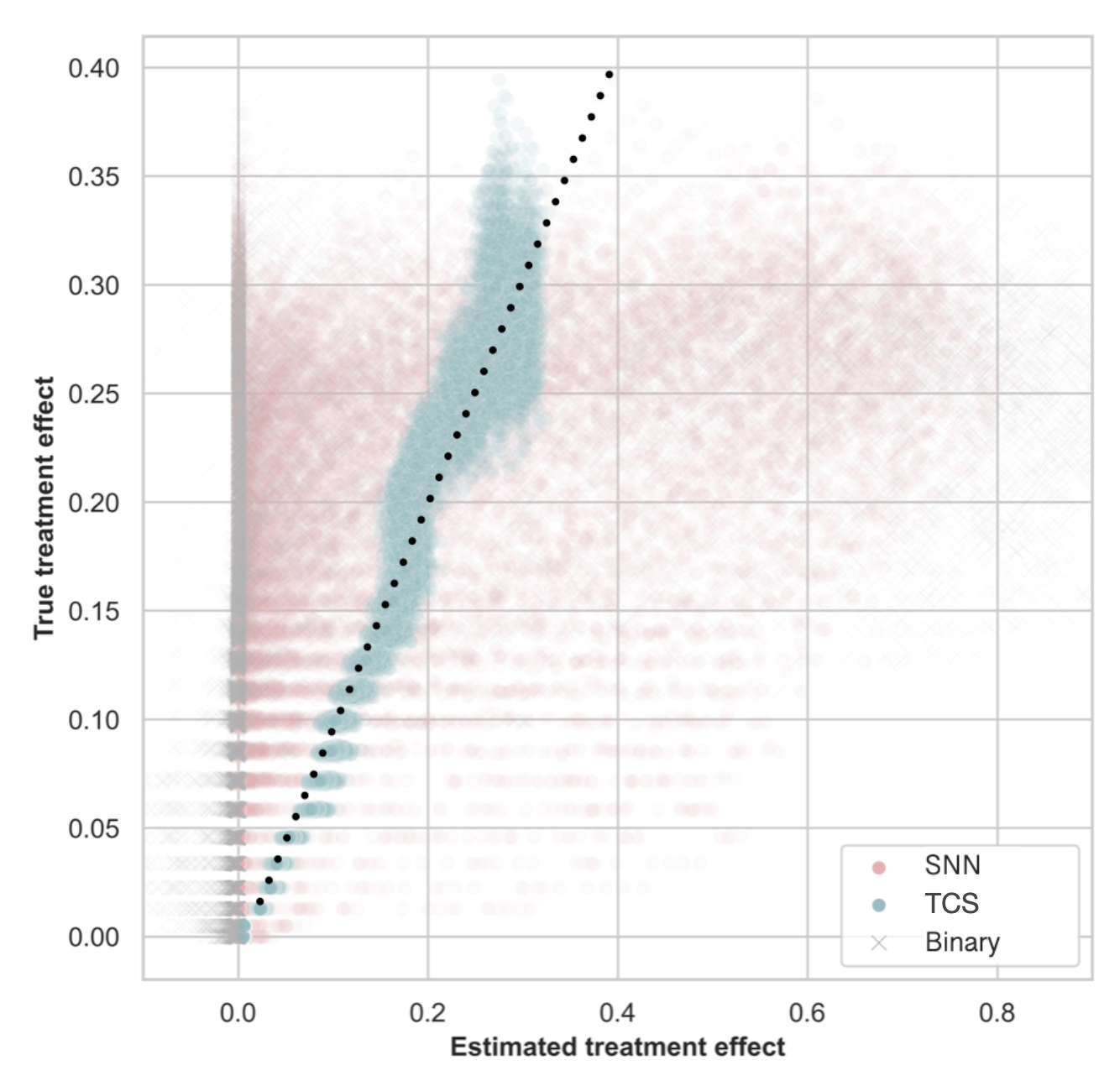}}{}
\caption{Diagnostic plots for the TCS model.
True and estimated individual treatment effect (ITE) distributions by benchmark algorithms. The colored dots are the true and estimated ITE averaged over of a randomly chosen sample under the default scenario. The dashed diagonal line indicates the equation of the true and estimated values.}
\label{f-5c8e52697469}
\end{figure*}

The improvement of TCS is most noticeable in the estimation of the ITEs, where the Bias is only 0.10 (0.064,0.137) across 50 samples compared to 0.43 (0.401,0.459) of the SNN model. However, the improvement of ATE estimation by TCS compared to TMLE or IPW adjusted SNN estimation is less significant, at around 5\%. TCS gains from its design of the propensity score layer as well as the potential outcomes subnetworks. In Figure~\ref{f-5c8e52697469}, we illustrate how TCS provides ITE estimations close to the true values, unlike the Binary and the SNN models. In particular, the Binary model only maximises its discrimination performance in terms of AUROC but provides no value to the treatment effect estimation.

\begin{table}[!ht]
\centering
\begin{threeparttable}
\footnotesize
\caption{Simulation study results.}
\label{f-result}
\begin{tabular}{lllll}
\hline
& \textbf{Bias (ATE)}& \textbf{Coverage} & \textbf{Bias (ITE)}& \textbf{RMSE}\\
\hline
\textbf{Overlap ($\eta$)} & \multicolumn{1}{c}{} & \multicolumn{1}{c}{} & \multicolumn{1}{c}{} & \multicolumn{1}{c}{} \\
\textbf{0.7} & 0.11 (0.070,0.157) & 0.70 (0.644,0.756) & 0.12 (0.081,0.167) & 1.04 (0.630,1.420) \\
\textbf{0.8} & 0.11 (0.063,0.151) & 0.72 (0.618,0.824) & 0.11 (0.063,0.151) & 1.08 (0.367,1.801) \\
\rowcolor[HTML]{ecebeb} 
\textbf{0.9} & 0.10 (0.061,0.136) & 0.90 (0.804,0.995) & 0.10 (0.064,0.137) & 1.04 (0.330,1.743) \\
\textbf{1} & 0.06 (0.012,0.107) & 0.98 (0.953,1.000) & 0.06 (0.014,0.106) & 0.56 (0.119,1.003) \\
\hline
\textbf{Dimension (D)} &&&&\\
\rowcolor[HTML]{ecebeb} 
\textbf{6} & 0.10 (0.061,0.136) & 0.90 (0.804,0.995)& 0.10 (0.064,0.137) & 1.04 (0.330,1.743) \\
\textbf{10}& 0.09 (0.050,0.135) & 0.94 (0.863,1.000)& 0.08 (0.042,0.125) & 1.05 (0.547,1.553) \\
\textbf{20}& 0.10 (0.044,0.161) & 0.94 (0.906,0.967)& 0.10 (0.044,0.161) & 1.29 (0.518,2.061) \\
\textbf{40}& 0.08 (0.022,0.130) & 0.93 (0.885,0.971)& 0.08 (0.023,0.130) & 1.09 (0.627,1.552) \\
\hline
\textbf{Variance (V)} &&&&\\
\rowcolor[HTML]{ecebeb} 
\textbf{0.5} & 0.10 (0.061,0.136) & 0.90 (0.804,0.995) & 0.10 (0.064,0.137) & 1.04 (0.330,1.743) \\
\textbf{1.0}& 0.10 (0.062,0.143) & 0.96 (0.924,1.004) & 0.10 (0.066,0.143) & 1.02 (0.661,1.379) \\
\textbf{1.5} & 0.11 (0.063,0.166) & 0.83 (0.746,0.924) & 0.12 (0.066,0.166) & 1.21 (0.513,1.914) \\
\textbf{2.0} & 0.08 (0.020,0.135) & 0.87 (0.794,0.949) & 0.10 (0.038,0.155) & 1.10 (0.490,1.715)\\
\hline
\textbf{Time}&& & &\\
\rowcolor[HTML]{ecebeb} 
\textbf{1} & 0.10 (0.101,0.101) & 0.88 (0.875,0.875) & 0.10 (0.101,0.101) & 0.10 (0.101,0.101) \\
\rowcolor[HTML]{ecebeb} 
\textbf{5} & 0.09 (0.053,0.137) & 0.94 (0.891,0.993) & 0.10 (0.054,0.138) & 0.94 (0.130,1.744) \\
\rowcolor[HTML]{ecebeb} 
\textbf{10} & 0.10 (0.059,0.139) & 0.89 (0.802,0.983) & 0.10 (0.061,0.139) & 1.16 (0.397,1.922) \\
\bottomrule
\end{tabular}
\begin{tablenotes}
\footnotesize 
\item{All metrics are averaged across 50 independent simulations over 30 time points from the test dataset under the default scenario. The shaded row indicates the default scenario.}
\end{tablenotes}
\end{threeparttable}
\end{table}

The performance of TCS over time is examined across different scenarios in Table~\ref{f-result}. The performance of TCS stands out when there is perfect overlapping ($\eta = 1$). In this case, the bias (as well as RMSE) of ATE and ITE estimations are about half as that in the decreased overlapping scenarios. Similarly, coverage is close to perfect when $\eta = 1$, at 0.98 (0.953,1.005). As the degree of overlapping drops, there is no significant difference in estimation accuracy in terms of the Bias, but the coverage rate declines dramatically from 0.90 (0.804,0.995) when $\eta = 0.9$ to 0.70 (0.644,0.756) when $\eta = 0.7$ due to decreased confidence in individual estimations. TCS is stable over scenarios with different sample dimensions from the default 6 confounders to the high-dimension 40-confounder scenario. However, the estimation accuracy declines with higher sample variance, when the sample variance is high ($V=2.0$), the coverage declined by about 3\% to 0.87 (0.794,0.949) compared to the default scenario. Lastly, we found the confidence interval widens over time, but there is no deterioration in the effect estimation accuracy. 

In Appendix C, we repeated the above scenarios with two additional sample sizes $N = 3000$ and $N=10000$. We found larger sample sizes improve the estimation accuracy for more complex data (i.e., higher level of dimension and sample variance). However, they can not help to improve the estimation for samples that lack overlapping. For observational EHR data, using a sample with moderate to high levels of overlapping is necessary to achieve better estimation accuracy.

\subsection{Treatment effect estimation with clinical data}

\begin{figure*}[!ht]
\centering\IfFileExists {images/case.png}{\includegraphics[width=.9\linewidth] {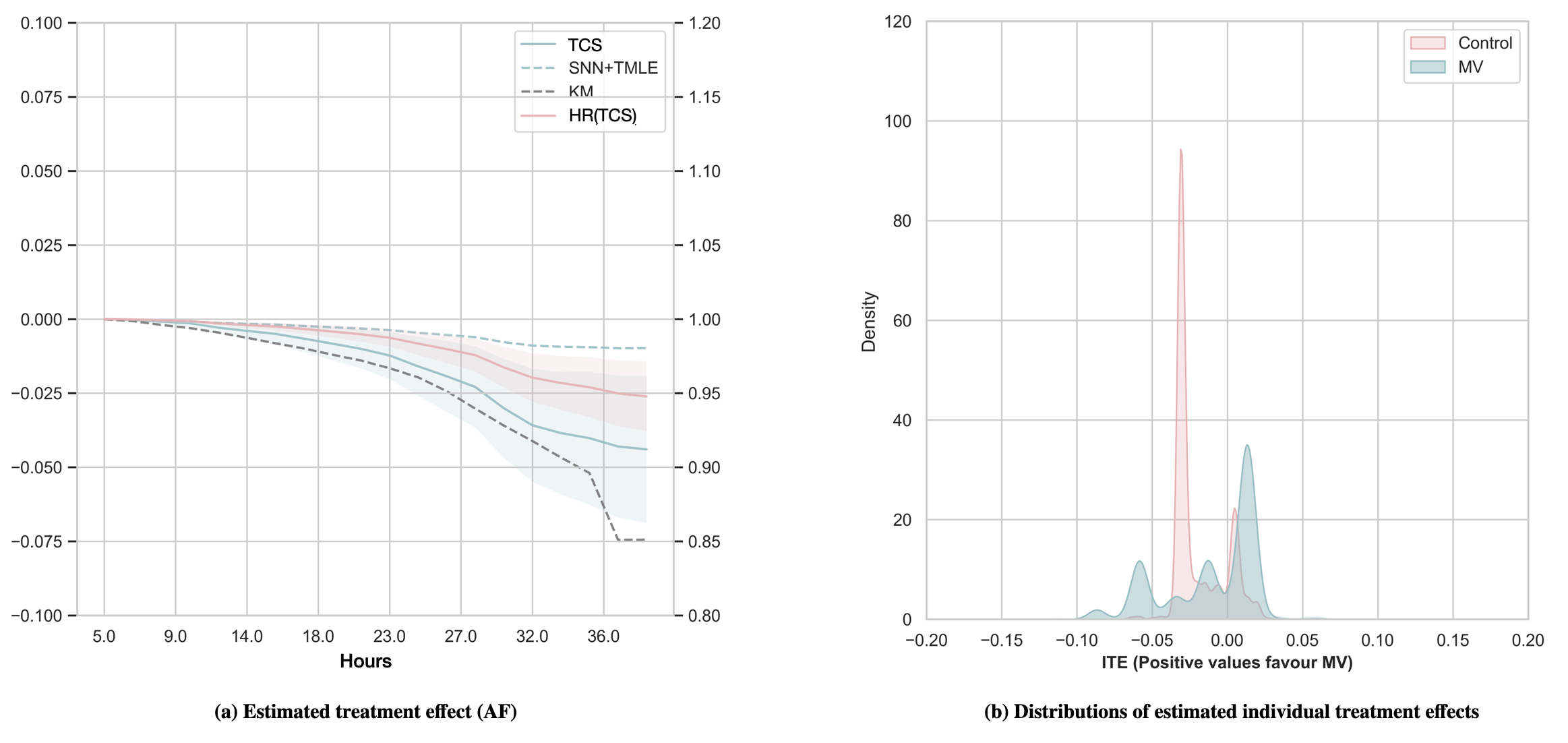}}{}
\caption{(a) Estimated treatment effect and empirical hazard ratio (HR$^*$) by benchmark algorithms. The shaded area is the 95\% confidence interval of the individual treatment effect estimations. (b) Distributions of estimated individual treatment effects (ITE) averaged over follow-up time. ITE estimations averaged over time in each study. Abbreviations: MV: mechanical ventilation; KM: the Kaplan Meier estimator; TMLE: the SNN model with TMLE adjustment; Hazard Ratio*: the empirical hazard ratio.}
\label{f-case1}
\end{figure*}

Figure~\ref{f-case1} (a) shows the estimation of ATE in terms of the differences in survival probability using TCS and SNN with TMLE adjustment (labeled as SNN+TMLE). We compared this absolute measure with the inversed empirical hazard ratio (i.e., $1/\text{HR}^*$, this is to make the hazard ratio in the same direction as the absolute difference in survival curves) and found that both curves closely follow each other. 

Panel (a) depicts the effect of using mechanical ventilation (MV) on the mortality of sepsis patients. The first 12 hours of data were used to estimate the treatment effect of MV on patient mortality for the subsequent 40 hours. We found the empirical hazard ratio ranges from 0.947 to 0.999, suggesting a minimal impact of using mechanical ventilation. The estimation from TCS  indicates that the usage of MV has an increasing negative impact over time and by the end of the follow-up, we saw the MV is expected to increase the probability of death by up to 4.39\% (1.917\%, 6.873\%) using the estimation from TCS or 3.04\% (2.11\%, 6.54\%) using TMLE. 

However, the heterogeneity of the estimated treatment effect is salient. Figure~\ref{f-case1} (b) shows the distributions of ITEs averaged over time colored by the observed treatment conditions. We saw a negative average treatment effect of $-1.98\% (-3.066\%, -0.013\%)$ for patients in the control group, while a minimal positive effect of $1.04\% (-0.931\%, 3.018\%)$ for patients administrated with mechanical ventilation.

\section{Discussion}
We have developed a novel causal inference algorithm to estimate the individual potential survival response curves from time-variant observational data. It leverages information across individuals under different interventions with dedicated propensity layers and potential outcome subnetworks. We demonstrated significant gains in the accuracy of TCS over plain recurrent neural networks in estimating individual and conditional average treatment effects.

In addition to extensive simulations, we applied TCS to the MIMIC-III sepsis study. Compared to the standard neural networks for binary outcome predictions, we found TCS has similar performance for the estimation of survival curves as DeepHit and Super-Learner models (the estimation performance of survival curves is presented in our working paper \cite{zhu2021dynamic}), but it is superior in identifying the treatment effect heterogeneity over time than existing methods such as the Cox model. In particular, TCS estimates the causal effect by computing the adjusted potential survival curves under treatment and control conditions. The use of propensity scores as the input of the network helped to improve the estimation accuracy by correcting for confounding bias. In addition, TCS learns from the pattern of missing confounders of a time series using masking layers rather than imputing their values, and it efficiently captures the uncertainty of the estimation using an ensemble of networks with varied random seeds.

In this study, time-dependent patient history has been simulated to resemble the observed data from observational electronic health records. Since deep learning techniques do not assume specific functional dependencies of treatment or outcomes, we expect our simulation results to hold under other choices of functional forms. Nevertheless, more extensive testing of TCS under extreme scenarios such as low overlapping and the presence of unmeasured confounders are desired and left for future work.

Our estimation of confidence intervals for ITEs is conservative and results in a high coverage ratio. However, we observed the interval is responsive to the quality of input data. For scenarios with high levels of overlapping and low levels of sample variance, the corresponding confidence intervals are much narrower. Therefore, the usage of network ensembles effectively captures the model uncertainty.

When comparing the ATE estimation from TCS with that from traditional confounding adjustment methods, we found that using propensity scores as a regressor in the neural network can achieve similar if not better performance than TMLE and IPW. When looking at the estimation performance of ITEs, TCS significantly outperforms both methods.

Much of the challenge of longitudinal causal inference lays in its definition of treatment effect. In this study, the treatment effect for a given period $t=0$ to $t=q$ has been defined as the difference in survival probability given constant treatment vs. control throughout the follow-up window $t'-u$ to $ t'+q$. However, when we have $n$ treatments, we will face the choice of making the contrast among $n(n-1)/2$ pairs. Nonetheless, the solution becomes more complicated when we consider the effect of a switch in the treatment, that is, we need to consider the timing of the switch as well as the choice of effect contrast. Similarly, it is also arduous to analyze continuous treatments. Nonparametric methods have been proposed to either discretise treatment options \cite{powell2020quantile} or create splines to estimate the treatment effect on a single day \cite{kennedy2017nonparametric}. Few has been discussed for time-dependent variables or treatments. One study \cite{793296:20397006} proposed to use reinforcement learning to control the intravenous fluid dosage for sepsis patients, but it does not answer the question of causal effect nor does it adjust for potential confounding bias. It would be interesting for future studies to explore time-dependent deconfounded treatment recommendations.

The proposed model is limited in its ability to capture the bias arising from missing confounders or measurement errors, and cannot be reduced by collecting more data under the same experimental conditions. This is reflected in our scenario analysis where increasing the sample size cannot improve the estimation accuracy if the data lacks overlapping. With observational data, the impact of overlapping is often overlooked due to the limited ability to identify and collect potential confounders. A recent study \cite{wallach2020evaluation} found 74 out of 87 ($85.1\%$) articles on the impact of alcohol on ischemic heart disease risk spuriously ignored or eventually dismissed confounding in their conclusions. Albeit this study acknowledges the caveats when interpreting results from case studies, it will be important for future researches to quantify the aleatoric uncertainty for data-adaptive models. 

TCS fills the gap in causal inference using deep learning techniques for survival analysis. It considers time-dependent patient history. Its treatment effect estimation can be easily compared with conventional literature which uses relative measures of treatment effect. We expect TCS will be particularly useful for identifying and quantifying treatment effect heterogeneity over time under the ever complex observational health care environment. We expect to improve TCS in future works to further account for the feedback between covariates and the time-varying treatments.

\section*{Acknowledgment}
This work was supported by National Health and Medical Research Council, project grant no. 1125414.

\appendix 

\section{Model implementation}
\begin{figure*}[!htbp]
\centering 
\includegraphics[width = 0.55\linewidth]{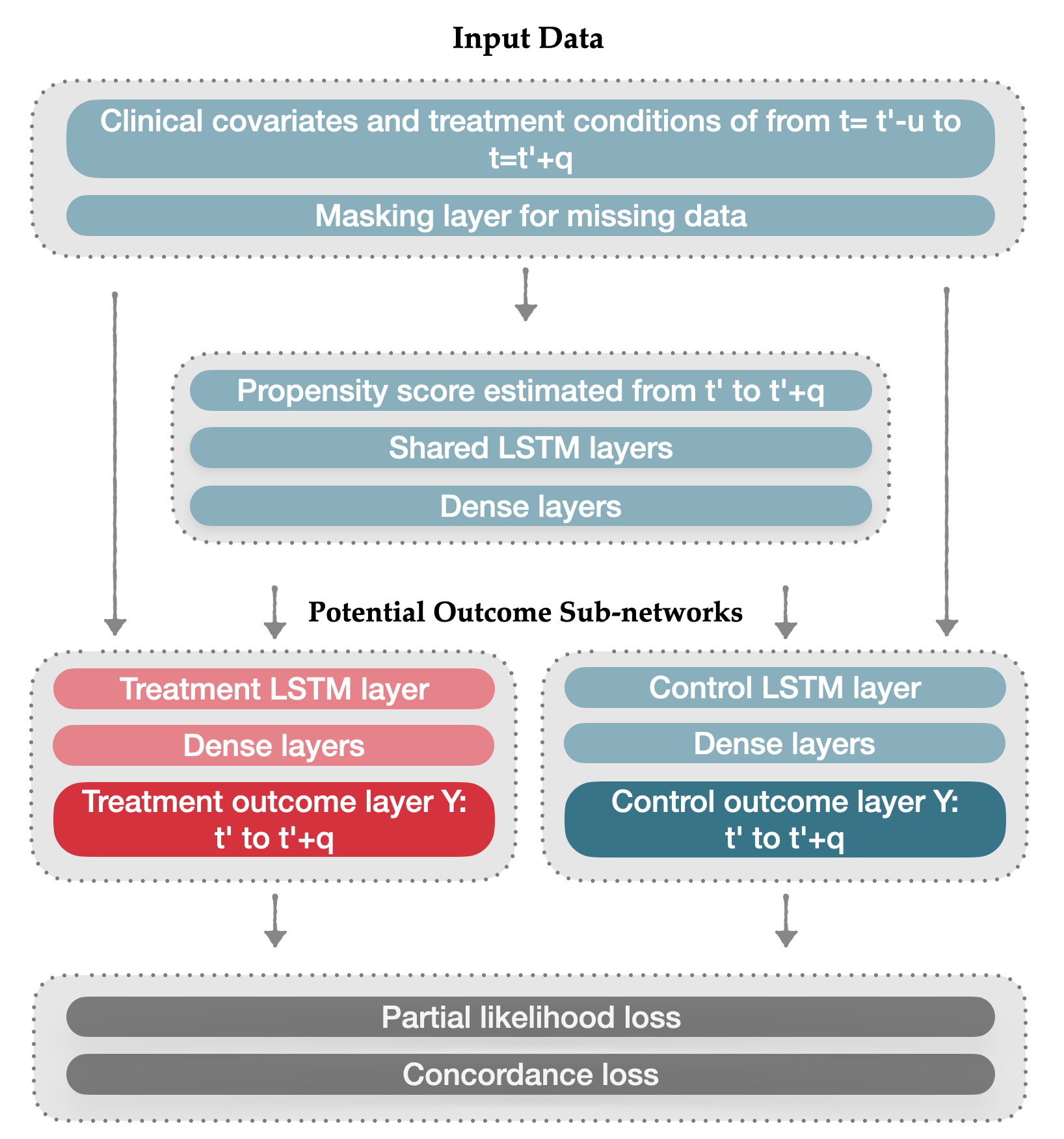}
\caption{Illustration of the TCS model. The input data is organized into a stacked matrix of covariates (number of observations (N) x covariate dimension (D) x [baseline steps (u) + follow-up steps (q)]). All missing data for future observations are masked. Abbreviations: $t'=0$, the start of follow-up; $t'+q$, the end of follow-up; $t'-u$, the start of patient history. Abbreviation: LSTM-Long short-term memory.  }
\label{f-aa387bd6e319}
\end{figure*}

The architecture of TCS model is illustrated in Figure \ref{f-aa387bd6e319}. This stacked matrix is first used to estimate the probability of treatment assignment via a densely connected neural network with Long Short-Term Memory (LSTM) units \cite{793296:19023733}. The output of this network is a vector of propensity scores. For each time point $t$ between $t=0$ and $t=q$, the propensity score of receiving treatment $A=a$ is given by:
\begin{align}
p(a(t)|\overline x(t)) :=\;Pr(A(t) = a(t)|\overline X(t) = \overline x(t)),
\end{align}
\noindent where $\overline{x}(t)$ is the history of covariates from $-u$ to $t$ (inclusive). In what follows, we denote $p(a(t)|\overline x(t))$ as $p(t)$ for simplicity.

We append the sequence of estimated propensities to the stacked matrix of $A(t), X(t)$ and get: 

\begin{align}
\begin{split}
\label{dfg-fefeqw3}
\Lambda & =\begin{bmatrix} 
        p(-1)& x(-u)& x(-u+1)&\ldots&x(-1)& x'(0)& x'(1)&\ldots&x'(q-1)& a(-1)   \\ 
        p(0)& x(-u)&x(-u+1)&\ldots&x(-1)& x(0)& x'(1)& \ldots&x'(q-1)& a(0)   \\ 
        p(\cdot)&x(-u)&x(-u+1)&\ldots&x(-1)&x(0)&\ldots& \ldots&x'(q-1)& a(\cdot) \\
        \ldots&\ldots&\ldots&\ldots&\ldots&\ldots&\ldots&\ldots&\ldots&\ldots  \\
        p(q-1)& x(-u)&x(-u+1)&\ldots&x(-1)& x(0)& x(1)& \ldots&x(q-1)& a(q-1)   \\ 
  \end{bmatrix},\\
\end{split}
\end{align}

\noindent where the prime symbol indicates the data is missing. 

The TCS then maps the input $\Lambda$ to the outcome $\Gamma$:

\begin{align}
 \Gamma = f(\Lambda)
\end{align}

The potential outcomes under the treatment condition are computed mapping $\hat\Gamma_1 = f(\Lambda_1)$, where $\Lambda_1$ is calculated by setting all $a(t)=1$ in $\Lambda$. Similarly, the potential outcomes under the control condition are computed mapping $\hat\Gamma_0 = f(\Lambda_0)$, where where $\Lambda_0$ is calculated setting all $a(t)=0$.

To train our neural network, we vectorized each individual event/censoring time to construct the target outcome in Equation~(\ref{dfg-364b5b929424}) and apply a loss function with two components: 

\noindent\textbf{1)} \textbf{the partial log likelihood loss}: the log likelihood loss of the joint distribution on the first event and censoring time:

\begin{align}
{\mathcal L}_{1} = \sum_{i}\bigg[\text{ln}(\gamma_i\hat{\Theta}_{i})+\text{ln}((1-\gamma_i) \hat{\Theta}_{i})\bigg]
\end{align}

\noindent which is a vector representation of the ordinary partial log-likelihood loss:

\begin{align}
&{\mathcal L}_{\text{uncensored}} = \sum_{i}e_i\bigg[\text{ln}(\hat\theta_i(t_i))+\sum_{j=0}^{t_i-1}\text{ln}(1-\hat\theta_i(j))\bigg]\\
&{\mathcal L}_{\text{censored}} = \sum_{i}(1-e_i)\bigg[\sum_{j=0}^{t_i-1}\big[1-\hat\theta_i(j)\big]\bigg],
\end{align}

\noindent  where $e_i=1$ if any $Y_i(t) = 1$  and $e_i=0$ if all $Y_i(t) = 0$ for all $t$. 

Therefore, each element $\hat\theta$ in the estimation $\hat{\Theta}$ is the conditional hazard rate, which is the probability of experiencing an event in interval $(t-1,t]$:
\begin{align}
       \hat\theta(t,\Gamma) :=  Pr(Y(t)=1 \vert\Gamma).
\end{align}

\noindent Then ${\mathcal L}_{1}$ can be written as:
\begin{align}
{\mathcal L}_{1} &= {\mathcal L}_{\text{uncensored}}+{\mathcal L}_{\text{censored}}\\
 &= \sum_{i}\bigg[e_i \text{ln}(\hat\theta_i(t_i))+e_i\sum_{j=0}^{t_i-1}\text{ln}(1-\hat\theta_i(j))+(1-e_i)\sum_{j=0}^{t_i-1}\text{ln}(1-\hat\theta_i(j))\bigg]\\
 &= \sum_{i}\bigg[e_i \text{ln}(\hat\theta_i(t_i))+\sum_{j=0}^{t_i-1}\text{ln}(1-\hat\theta_i(j))\bigg]
\end{align}

\noindent\textbf{2) the rank loss function}: the loss function associated with the concordance index in survival analysis \cite{793296:20320158}: a subject who experienced an event at time $t$ should have a higher probability of failure than a subject who doesn't or who is censored. We count the number of acceptable pairs of estimated hazard rate $\{\hat{\theta}_i(t),\hat{\theta}_j(t)\}$ in the loss function: 

\begin{eqnarray*}
{\mathcal L}_{2} =\sum_{t} \sum_{i \neq j} I_{ij} 
\end{eqnarray*}
where $I_{ij}$ is an indicator function: 
\begin{eqnarray*}
I_{ij} = 1, \text{if } Y_i(t) = 1, Y_j(t) = 0 \text{ and } \hat{\theta}_i(t) > \hat{\theta}_j(t),
\end{eqnarray*}
$I_{ij} = 0$, otherwise. 

The final loss function is defined as:

\begin{eqnarray*}
\begin{array}{@{}l}{\mathcal L} = \alpha{\mathcal L}_{1} + \beta{\mathcal L}_{2}, \end{array}
\end{eqnarray*}
\noindent where random search is used to locate the best hyper-parameters $\alpha \text{ and } \beta$. To capture the uncertainty of the neural network, we iterate the model training with different random seeds for $20$ iterations and average the results.

The estimated probability from $\hat\Theta$ is therefore the hazard rate adjusted for the probability of right censoring \cite{793296:19503682}. Following from our previous work \cite{zhu2020targeted}, the probability that an individual will experience an event after time $t$ can be written as a product of 'hazard functions describing the conditional probability that the event did not occur in any observation:
\begin{align}
\begin{split}
\label{dfg-2c1cf113437b}
S(t,\Lambda) &= \prod_{j=1}^{t}(1-\hat\theta(j,\Lambda)).
\end{split}
\end{align}

we use $S(t,\Lambda_1)$ to denote the time-to-event probabilities given patient receives treatment throughout the follow-up period and $S(t,\Lambda_0)$ to denote the time-to-event probabilities given patient receives the control/comparator intervention throughout the follow-up period.

\section{Average treatment effect estimation adjustment}

\subsection{Inverse probability weighting (IPW)}

We apply the inverse probability weighting adjustment to the raw estimation of ATE with the following equation:
\begin{eqnarray*}\hat\psi_{IPW}(t) =\frac{1}{N}\sum_{i=1}^{n} \bigg( \frac{A_i \hat Y_i(t)}{\hat P(X_i(0))} - \frac{(1-A_i) \hat Y_i(t)}{1-\hat P(X_i(0))} \bigg), \text{ where }t\in \{0,1,\ldots,q\} \end{eqnarray*}
where $N $ is the sample size, $q $ is the maximum of follow-up time and $\hat P(X_i(0)) $ is the propensity score estimated as the probability of receiving the treatment at time 0 if the treatment assignment is time-invariant. When the treatment is time-variant, we estimate the propensity score at each time step as $\hat P(X_i(t)) $. In this study, we estimated $\hat P(X_i(0)) $ using a densely connected network to fit the binary label of the treatment assignment of each subject $i $ at time 0. 

\subsection{The iterative targeted maximum likelihood estimation (TMLE)}

To apply the iterative targeted maximum likelihood estimation adjustment, we conducted the following adjustment at each time step:

1. We first calculate the smart covariates $H(A,X(t)) $ using the propensity score estimated using the procedure aforementioned: 
\begin{eqnarray*}H(1,X_i(t)) = \frac{A_i }{\hat P(X_i(0))}; H(0,X_i(t)) = \frac{1-A_i }{1-\hat P(X_i(0))} \end{eqnarray*}
2. Then we fit the residual of the initial estimate of the logit of the binary label with smart covariates using an intercept-free regression: 
\begin{eqnarray*}logit(\hat Y_i(t))-logit(Y_i(t))=\delta_1(t) h(1,X_i(t)) +\delta_0(t) h(0,X_i(t))
 \end{eqnarray*}
where $logit(x) $ represents the function $log(\frac{x}{1-x}) $

3. Calculate the adjusted potential outcomes:
\begin{eqnarray*}
\hat Y_A(t)^{1} = log \bigg(\frac{logit(\hat Y_A(t)+ \delta_A(t)}{P_A(X_i(0))}\bigg), \text{ for } A \in \{0,1\} \end{eqnarray*}
where $\hat P_1(X_i(0)) = \hat P(X_i(0)) $ and $\hat P_0(X_i(0)) = 1-\hat P(X_i(0)) $. 

4. Targeted estimate of ATE at time t:
\begin{eqnarray*}{\widehat\psi}_{TMLE}(t)=\frac1N\sum_{i=1}^{n} (\hat Y_1(t)^{1}-\hat Y_0(t)^{1}) \end{eqnarray*}

\section{The structure of the TCS masking and outcome layers}

TCS estimates the the difference between potential survival curves under the treatment and control conditions to compute the estimated individual treatment effect (ITE) curve. Here, following the notations in the main manuscript, we describe the masking and outcome layers of the TCS model introduced in Figure \ref{f-aa387bd6e319} as follows:

\begin{itemize}
\item A masking layer taking account of informative missingness in longitudinal data \cite{793296:19020959}, which consists of two representations of missing patterns, i.e., a masking vector $M_t\in \{0, 1\}^{D} $to denote which variables are missing at time $t $, and a real vector $\delta_t\in \mathbb{R}^{D,q} $ to capture the time interval for each variable $d $ since its last observation over $q $time points. The masking layer takes as inputs the matrix $[\overline{X}(u), \overline{A}(u)]$ and produces as output a matrix $[\overline{M}(u),\overline\delta(u),\overline{X}(u),\overline{A}(u)] $, where the overlines indicate the corresponding vector observed during the history window $[t-u, t-1] $. This layer effectively uses the missing data patterns to achieve better predictions; 
 
\item The potential outcome layers make predictions of the log odds of the binary outcomes given by $Y^M$ given$[\overline{M}(u),\overline\delta(u),\overline{X}(u),\overline{A}(u)=1] $ and $[\overline{M}(u),\overline\delta(u),\overline{X}(u),\overline{A}(u)=0]$ and then convert the log odds into the conditional survival probability to form the potential survival curves under each treatment condition. 
\end{itemize}

\section{Additional simulation results}

\begin{figure*}[!hbp]
\centering 
\includegraphics[width = 1.00\linewidth] {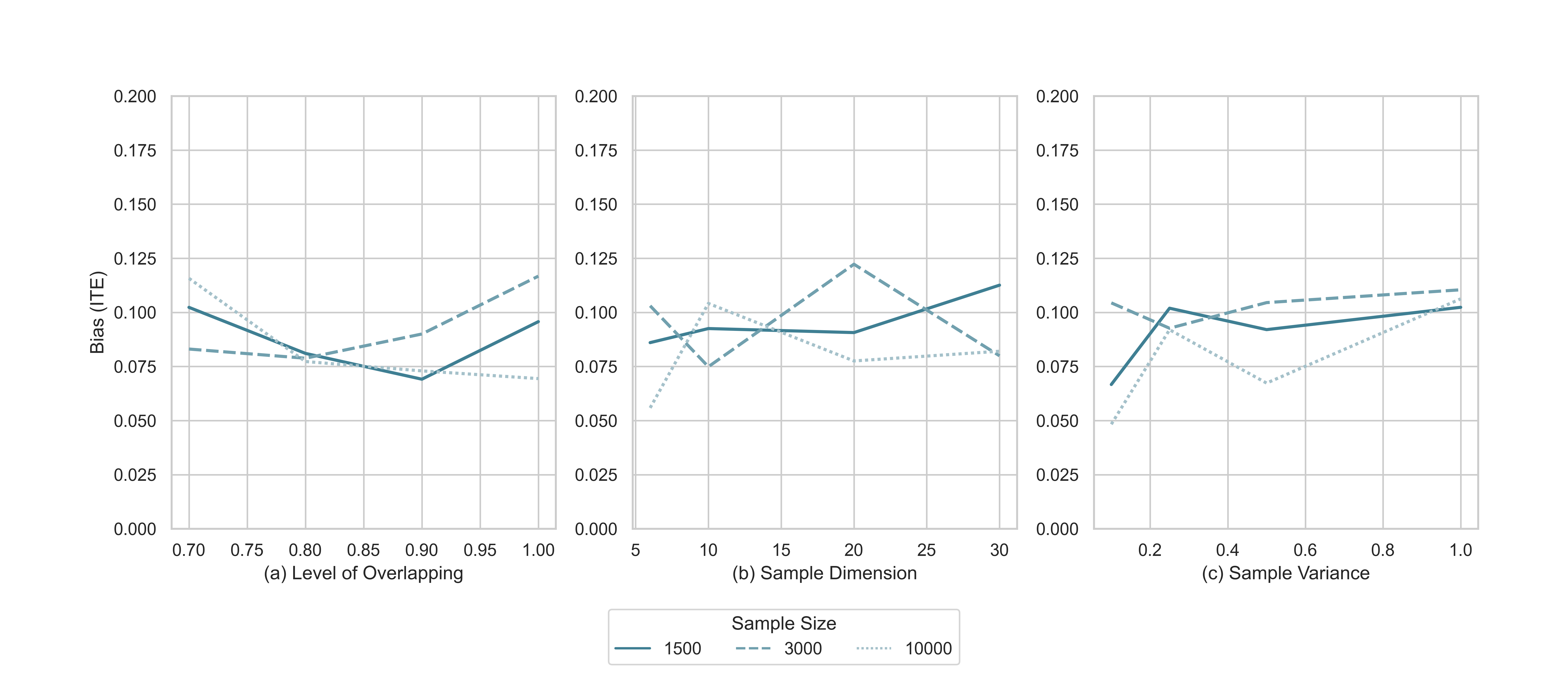}
\caption{Individual treatment effect (ITE) estimation bias by scenarios. All metrics are averaged across 50 independent simulations over 10 time points from the test dataset under each scenario. Bias: average absolute percentage bias.}
\label{f-44d629814639}
\end{figure*}

\clearpage

\section{Descriptive statistics for empirical databases}

\begin{table}[h]
\centering
\footnotesize
\caption{Descriptive statistics for case study 2}
\label{case2}
\renewcommand{\arraystretch}{1.1}
\setlength\tabcolsep{4pt}
\begin{tabular}{lrllll}
\textbf{} & \textbf{Count} & \multicolumn{1}{r}{\textbf{Mean}} & \multicolumn{1}{r}{\textbf{SD}} & \multicolumn{1}{r}{\textbf{0.25}} & \multicolumn{1}{r}{\textbf{0.75}} \\
\hline
\rowcolor[HTML]{F2F2F2} 
Unique ID & 6225 &  &  &  &  \\
Rows & 98716 &  &  &  &  \\
\rowcolor[HTML]{F2F2F2} 
\textbf{Death (1 = Yes, 0 = No)} & 459 (7.4\%) &  &  &  &  \\
\textbf{Vesopressor Dosage (µg/kg/min)} &  & 0.29 & 1.913 & 0.00 & 0.14 \\
\rowcolor[HTML]{F2F2F2} 
Follow-Up Hours &  & 35.16 & 20.354 & 16.00 & 52.00 \\
Surgery & 350 (5.6\%) &  &  &  &  \\
\rowcolor[HTML]{F2F2F2} 
Age &  & 64.72 & 14.074 & 57.00 & 75.00 \\
Gender (1 = Male, 0 = Female) & 3647 (58.6\%) &  &  &  &  \\
\rowcolor[HTML]{F2F2F2} 
Glasgow Coma Scale (GCS) &  & 169.35 & 16.448 & 162.60 & 177.80 \\
Heart Rate (Bp/S) &  & 83.38 & 28.676 & 73.15 & 100.13 \\
\rowcolor[HTML]{F2F2F2} 
Spo2 (\%) &  & 89.41 & 25.472 & 94.44 & 98.93 \\
Respiratory Rate (Breaths/Min) &  & 18.24 & 8.163 & 15.04 & 22.85 \\
\rowcolor[HTML]{F2F2F2} 
Non-Invasive BP Systolic (Mmhg) &  & 84.63 & 51.103 & 62.25 & 117.88 \\
Non-Invasive BP Diastolic (Mmhg) &  & 46.17 & 28.420 & 28.99 & 65.25 \\
\rowcolor[HTML]{F2F2F2} 
Non-Invasive BP Mean (Mmhg) &  & 56.16 & 34.852 & 0.00 & 79.25 \\
Temperature (Celsius) &  & 27.61 & 16.120 & 0.00 & 37.20 \\
\rowcolor[HTML]{F2F2F2} 
Shock Index &  & 0.61 & 0.418 & 0.00 & 0.88 \\
Sodium (Mmol/L) &  & 43.46 & 64.717 & 0.00 & 135.00 \\
\rowcolor[HTML]{F2F2F2} 
Potassium (Mmol/L) &  & 1.44 & 2.016 & 0.00 & 3.70 \\
Chloride (Mmol/L) &  & 31.21 & 48.566 & 0.00 & 99.00 \\
\rowcolor[HTML]{F2F2F2} 
Glucose (Mg/Dl) &  & 45.94 & 81.823 & 0.00 & 101.00 \\
Blood Urea Nitrogen (BUN, Mg/Dl) &  & 9.41 & 19.110 & 0.00 & 13.00 \\
\rowcolor[HTML]{F2F2F2} 
Creatinine (Mg/Dl) &  & 0.55 & 1.206 & 0.00 & 0.73 \\
Magnesium (Mg/Dl) &  & 0.42 & 0.859 & 0.00 & 0.00 \\
\rowcolor[HTML]{F2F2F2} 
Calcium (Mg/Dl) &  & 2.23 & 3.592 & 0.00 & 6.90 \\
Total Bilirubin (Mg/Dl) &  & 0.21 & 1.206 & 0.00 & 0.00 \\
\rowcolor[HTML]{F2F2F2} 
AST (SGOT) (Units/L) &  & 62.77 & 643.250 & 0.00 & 0.00 \\
ALT (SGPT) (Units/L) &  & 35.97 & 319.727 & 0.00 & 0.00 \\
\rowcolor[HTML]{F2F2F2} 
Albumin (G/Dl) &  & 0.35 & 0.928 & 0.00 & 0.00 \\
Hgb (G/Dl) &  & 2.94 & 4.751 & 0.00 & 8.00 \\
\rowcolor[HTML]{F2F2F2} 
White Blood Cell Count (K/Mcl) &  & 3.67 & 7.867 & 0.00 & 2.00 \\
Platelets Count (K/Mcl) &  & 42.32 & 87.025 & 0.00 & 34.00 \\
\rowcolor[HTML]{F2F2F2} 
Partial Thromboplastin Time (PTT, Sec) &  & 4.69 & 16.312 & 0.00 & 0.00 \\
Prothrombin Time (PT,Sec) &  & 2.33 & 7.374 & 0.00 & 0.00 \\
\rowcolor[HTML]{F2F2F2} 
International Normalized Ratio (INR) &  & 0.23 & 0.739 & 0.00 & 0.00 \\
Arterial Ph &  & 1.92 & 3.228 & 0.00 & 7.13 \\
\rowcolor[HTML]{F2F2F2} 
Pao2 (Mmhg) &  & 30.49 & 62.572 & 0.00 & 54.00 \\
Paco2 (Mmhg) &  & 10.62 & 18.797 & 0.00 & 25.00 \\
\rowcolor[HTML]{F2F2F2} 
Base Excess (Meq/L) &  & -0.96 & 3.644 & 0.00 & 0.00 \\
Fio2 (\%) &  & 14.13 & 28.029 & 0.00 & 0.00 \\
\rowcolor[HTML]{F2F2F2} 
HCO3 (Mmol/L) &  & 5.27 & 9.702 & 0.00 & 0.00 \\
Lactate (Mmol/L) &  & 0.57 & 2.069 & 0.00 & 0.00 \\
\rowcolor[HTML]{F2F2F2} 
Pre-Admission Fluid Input (Ml) &  & 390.38 & 2657.466 & 0.00 & 0.00 \\
Pre-Admission Fluid Output (Ml) &  & 522.39 & 2513.307 & 0.00 & 100.00 \\
\rowcolor[HTML]{F2F2F2} 
Pre-Admission Balance (Ml) &  & -132.01 & 3072.695 & 0.00 & 0.00 \\
Fluid Input (Ml/4 Hours) &  & 54.88 & 316.422 & 0.00 & 0.00 \\
\rowcolor[HTML]{F2F2F2} 
Fluid Output (Ml/4 Hours) &  & 32.36 & 199.217 & 0.00 & 0.00 \\
Fluid Balance (Ml/4 Hours) &  & 22.52 & 328.079 & 0.00 & 0.00 \\
\hline
\end{tabular}%
\end{table}

\bibliographystyle{unsrt}
\bibliography{article.bib}
\end{document}